\pdfoutput=1
\documentclass[11pt]{article}
\usepackage{acl2023}

\usepackage{times}
\usepackage{latexsym}
\usepackage{hyperref}
\usepackage[utf8]{inputenc}
\usepackage[T1]{fontenc}
\usepackage{microtype}
\usepackage{enumitem} 
\newlist{compactenum}{enumerate}{4}
\setlist[compactenum,1]{nolistsep}
\usepackage{paralist}

\usepackage{graphicx}
\usepackage{longtable}
\usepackage{array}
\usepackage{booktabs}
\usepackage{subcaption}

\usepackage{inconsolata}
\usepackage{graphicx}
\usepackage{multirow}
\usepackage{caption}
\usepackage{float}
\usepackage[none]{hyphenat}

\usepackage{listings}
\usepackage{tikz}
\def\checkmark{\tikz\fill[scale=0.4](0,.35) -- (.25,0) -- (1,.7) -- (.25,.15) -- cycle;} 

\usepackage{amsmath}

\title{SEA-HELM: \\ Southeast Asian Holistic Evaluation of Language Models}

\author{
 \textbf{Yosephine Susanto\textsuperscript{1,2}},
 \textbf{Adithya Venkatadri Hulagadri\textsuperscript{1,2,*}},
 \textbf{Jann Railey Montalan\textsuperscript{1,2.*}},
\\
 \textbf{Jian Gang Ngui\textsuperscript{1,2,*}},
 \textbf{Xian Bin Yong\textsuperscript{1,2,*}},
 \textbf{Weiqi Leong\textsuperscript{1,2}},
 \textbf{Hamsawardhini Rengarajan\textsuperscript{1,2}},
\\
 \textbf{Peerat Limkonchotiwat\textsuperscript{1,2}},
 \textbf{Yifan Mai\textsuperscript{3}},
 \textbf{William Chandra Tjhi\textsuperscript{1,2}}
\\
\\
 \textsuperscript{1}AI Singapore,
 \textsuperscript{2}National University of Singapore
 \\
 \textsuperscript{3}Center for Research on Foundation Models (CRFM), Stanford University
\\
 \textsuperscript{*}Core contributors
\\
 \small{
   \textbf{Correspondence:} \href{mailto:email@domain}{yosephine@aisingapore.org}
 }
}

\begin{document}

\maketitle
\begin{abstract}

With the rapid emergence of novel capabilities in Large Language Models (LLMs), the need for rigorous multilingual and multicultural benchmarks that are integrated has become more pronounced. 
Though existing LLM benchmarks are capable of evaluating specific capabilities of LLMs in English as well as in various mid- to low-resource languages, including those in the Southeast Asian (SEA) region, a comprehensive and culturally representative evaluation suite for the SEA languages has not been developed thus far.
Here, we present \textbf{SEA-HELM},\footnote{Formerly known as BHASA \citep{leong2023bhasa}.} a holistic linguistic and cultural LLM evaluation suite that emphasises SEA languages, comprising five core pillars: (1) \textsc{NLP Classics}, (2) \textsc{LLM-specifics}, (3) \textsc{SEA Linguistics}, (4) \textsc{SEA Culture}, (5) \textsc{Safety}. 
SEA-HELM currently supports Filipino, Indonesian, Tamil, Thai, and Vietnamese. We also introduce the \textbf{SEA-HELM leaderboard},\footnote{\url{https://leaderboard.sea-lion.ai/}} which allows users to understand models' multilingual and multicultural performance in a systematic and user-friendly manner. We make the SEA-HELM evaluation code publicly available.\footnote{\url{https://github.com/aisingapore/SEA-HELM}}

\end{abstract}

\section{Introduction}
The proliferation of generative approaches to natural language processing (NLP) through Large Language Models (LLMs) has rendered many traditional datasets for NLP evaluation compromised \citep{haimes2024benchmark}, obsolete or saturated \citep{liu2024benchmarkinggenerationevaluationcapabilities}. While essentially trained to predict the next token in a sequence, LLMs have shown significant emergent competencies, including summarisation, question answering, translation, coding, and advanced reasoning \cite{brown_language_2020,yeo2024selftraining}. They are also extensively used for new applications, such as chatbots that can hold sustained open-ended conversations \citep{dam2024complete}. This advancement has led to a significant disparity between the range of LLM capabilities and the datasets and frameworks to evaluate them rigorously. Traditional approaches to NLP evaluation, which emphasise on alignment with a predefined ground truth reference, are not sufficient in measuring the complex abilities of LLMs \citep{van2024field, gema2025mmlu, grusky-2023-rogue}. This gap is further exacerbated in lower-resource languages in \textbf{S}outh\textbf{E}ast \textbf{A}sia (\textbf{SEA}), owing to a lack of both training and testing data on the internet \citep{li2024languagerankermetricquantifying}. 

\begin{figure}[t]
    \centering
    \includegraphics[width=1\linewidth]{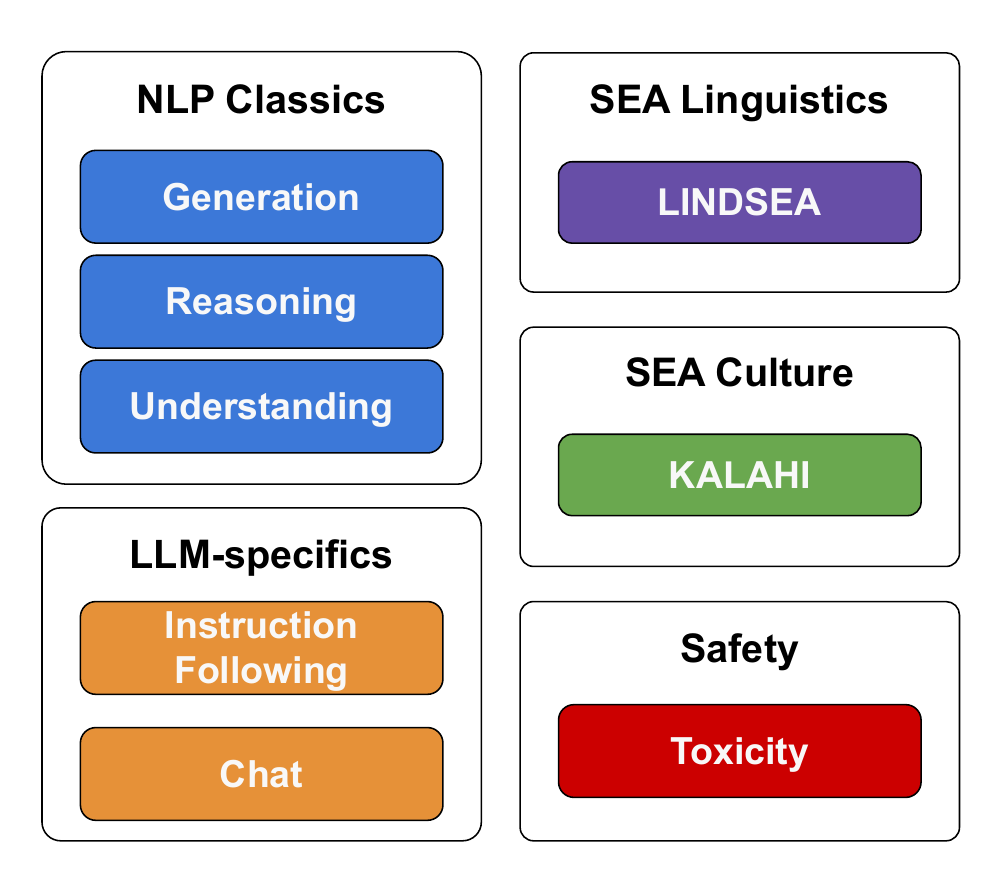}
    \caption{The five evaluation pillars of SEA-HELM that make up our holistic and integrated approach.} 
    \label{fig:sea-helm}
\end{figure}

The SEA region is home to nearly 700 million speakers across more than 1,000 languages (see Appendix \ref{tab:sea_population}). While this region represents almost 10\% of the global population and constitutes approximately one-seventh of the world's total languages, most of these languages remain unsupported by major LLMs, such as Mistral \citep{mistral2023}  and Claude \citep{anthropic2023claude}. The complications arising from a lack of data as well as uneven digital access and representation contribute to the impeded development of LLMs in those languages in the SEA region. This matter is further aggravated by the fact that some SEA languages are written in non-Latin scripts, which presents a challenge for tokenizers when processing limited data.

Despite the mentioned obstacles, multilingual LLM and benchmark development in SEA strive to close the gap and adapt to the current trends in the field. Some models now explicitly support SEA languages and also claim to provide representation for SEA cultural knowledge \citep{sailor2report,zhang2024seallms,bai2023qwen,dang2024aya}. There have also been many benchmarks claiming to measure LLMs' multilingual and multicultural capabilities for the SEA region \citep{damonlpsg2023seallm,damonlp2024seallm3,wang2023seaeval,SeaExam2023,singh2024global,CohereForAI_mArenaHard,lovenia2024seacrowdmultilingualmultimodaldata}. However, there has yet to be a comprehensive and culturally representative benchmark suite for evaluating LLMs for SEA cultural and linguistic competencies.

We have thus developed \textbf{SEA-HELM} (\textbf{S}outh\textbf{E}ast \textbf{A}sian \textbf{H}olistic \textbf{E}valuation of \textbf{L}anguage \textbf{M}odels),\footnote{We release SEA-IFEval, SEA-MTBench and LINDSEA datasets under the Creative Commons Attribution Share-Alike 4.0 (CC-BY-SA 4.0) license. This respects the licenses of the source datasets used in this study.} 
a systematic, integrated, and continually maintained benchmark suite which aims to measure the SEA language and cultural competencies of LLMs in a targeted and comprehensive manner. SEA-HELM achieves integration by collating localised evaluation datasets and LLM prompts, running tests on models together to enable standardised comparisons, and presenting results aggregated by languages, tasks and models. It is our view that no single metric explains a model's suitability for SEA, and thus SEA-HELM is designed to test a holistic set of competencies, illustrated in Figure \ref{fig:sea-helm}. 

Specifically, SEA-HELM is organised into five evaluation pillars: (1) \textsc{NLP Classics}; (2) \textsc{LLM-specifics}; (3) \textsc{SEA Linguistics}; (4) \textsc{SEA Culture}; and (5) \textsc{Safety}, which together encompass an extensive range of tasks for each SEA language to ensure that a wide range of relevant aspects, from linguistic nuances to cultural representations, are considered and accounted for. The five pillars are also meticulously and rigorously crafted to achieve fair, transparent, and authentic multilingual and multicultural evaluations of LLMs in the region. We deliberately incorporate community participation by involving native speakers of the SEA languages at each stage of the dataset planning and construction to ensure linguistic accuracy and cultural authenticity.

We summarise the contributions of \textbf{SEA-HELM} here:

\begin{compactitem}[\hspace{3pt}•]
    \item SEA-HELM is a curated suite of SEA datasets which are evaluated together and whose results will be presented on a publicly visible leaderboard. 
    \item SEA-HELM (a) incorporates natively adapted and human-translated benchmark tasks from a range of SEA languages such as Filipino,\footnote{Which also culminated in \textsc{Batayan}, the Filipino expansion of SEA-HELM.} Indonesian, Tamil,\footnote{Tamil is one of the official languages of Singapore and is also spoken in Malaysia.} Thai and Vietnamese; (b) created the human-translated SEA-IFEval and SEA-MTBench datasets; and (c) constructed LLM prompt templates in each native language that are consistent across tasks.
    \item SEA-HELM includes new datasets we developed for granular linguistic diagnostics (LINDSEA) in Indonesian and Tamil. SEA-HELM also includes a cultural evaluation dataset for Filipino, \textsc{Kalahi}, developed in collaboration with community members from the Philippines \citep{montalan2024kalahihandcraftedgrassrootscultural}.
\end{compactitem}

We believe that these make SEA-HELM a comprehensive and representative evaluation suite for SEA languages. It can be used as a base for future extensions covering other SEA languages, e.g. Khmer, Lao, Burmese, etc., which we intend to expand on in a future work.

\section{Related work} 
\subsection{LLM evaluations}
Over the years, AI practitioners have employed either an individual task-based or, more rarely, a holistic approach to assess the performance and capabilities of LLMs. Popular tasks for evaluating LLMs include translation \cite{hendy_how_2023}, summarisation \cite{zhang_benchmarking_2023}, decision-making \citep{shen_hugginggpt_2023}, detecting scalar implicatures \citep{jeretic_are_2020,pandia_pragmatic_2021,hu_expectations_2023, liu_adjective_2023} as well as presuppositions \citep{jeretic_are_2020,parrish_nope_2021}. Additionally, linguistic \citep{warstadt_blimp_2020, xiang_climp_2021,someya-oseki-2023-jblimp} and cultural representation \citep{durmus_towards_2023, atari2023humans} are also increasingly recognised as essential criteria for evaluating the efficacy and fairness of language models. 

On a holistic approach, Stanford University introduced HELM \citep{liang_holistic_2022} as an initiative aimed at evaluating LLMs across a wide range of tasks, such as linguistic capabilities, reasoning, knowledge, memorisation, disinformation, bias and toxicity. Google introduced BIG-Bench \citep{srivastava_beyond_2023}, which is a crowdsourced initiative. Similarly, OpenAI launched OpenAI Evals,\footnote{\url{https://github.com/openai/evals/}} a crowdsourced system that invites users to create custom evaluation datasets.

\subsection{LLM evaluations for SEA languages}

Recently, an increasing amount of attention has been directed towards LLM training and evaluations beyond English. There has been a growing body of work \citep{sailor2report,zhang2024seallms,bai2023qwen,dang2024aya} evaluating the performance of LLMs in a wide range of tasks in SEA languages. Most of them attempted to incorporate a broad spectrum of languages (e.g. Indonesian, Thai, Filipino). In order to achieve such large language coverage, machine translation and synthetic generation were typically used to generate multilingual benchmark datasets.

However, the use of machine translation and synthetically generated benchmarks with little input from the community raises questions on their cultural authenticity and reliability. Automatic translation often misses the cultural nuances inherent in the target language, and can result in translation errors and biases \citep{singh2024global}. This can result in cultural erasure, furthering stereotypical or non-diverse views \citep{qadri2025risksculturalerasurelarge}. Thus, there is a need to develop authentic, human-verified multilingual evaluation datasets and metrics. Works such as \citet{singh2024global}, \citet{romero2024cvqa}, and \citet{koto2024indoculture}
address the above point by adopting a participatory framework 
\citep{birhane2022power, smart2024socially}. The participatory framework is also core to SEA-HELM's design philosophy as it ensures linguistic accuracy and cultural authenticity.

\section{SEA-HELM}
To address the lack of holistic multilingual and multicultural evaluations for the SEA region, we designed and developed SEA-HELM, which draws its inspiration from HELM \citep{liang_holistic_2022}. This evaluation suite consists of five core pillars: (1) \textsc{NLP Classics}, (2) \textsc{LLM-specifics}, (3) \textsc{SEA Linguistics}, (4) \textsc{SEA Culture}, and (5) \textsc{Safety}, and has been recently integrated with HELM. The spread of tasks and languages is detailed in Table \ref{tab:dataset}. SEA-HELM currently supports five SEA languages -- Filipino, Indonesian, Tamil, Thai, and Vietnamese, enabling users and AI practitioners to assess the overall performance of LLMs for these languages.

\begin{table*}[t]
\centering
\resizebox{\textwidth}{!}{%
\begin{tabular}{p{2.5cm} p{2.5cm} p{3.5cm} lllllc}
\toprule
Pillar      & Competency & Task          & Dataset             & Language   & Metrics             & Native & Translation & Our contribution \\ \midrule
NLP Classics & NLU        & Sentiment     & PH Elections Sentiment \cite{cabasag2019hate}  & FIL   & WA & Y    &         -& \\
            &            &               & NusaX \cite{winata-etal-2023-nusax}              & ID & WA   & Y      &             -&              \\
            &            &               & IndicSentiment \cite{doddapaneni-etal-2023-towards}     & TA      & WA   & N      & Human       &              \\
            &            &               & Wisesight \cite{bact_2019_3457447}          & TH       & WA   & Y      &             -&              \\
            &            &               & UIT-VSFC \cite{nguyen_uit-vsfc_2018}           & VI & WA   & Y      &             -&             \\ \cmidrule{3-9} 
            &            & QA            & TyDi   QA-GoldP \cite{clark-etal-2020-tydi}    & ID & F1                  & Y      &             -&              \\
            &            &               & IndicQA \cite{doddapaneni-etal-2023-towards}            & TA  & F1                  & Y      &             -&              \\
            &            &               & XQUAD \cite{artetxe_cross-lingual_2020}              & TH, VI & F1                  & N      & Human       &              \\
            &            & QA-Multiple Choice & Belebele \cite{bandarkar-etal-2024-belebele} & FIL & F1   & N    & Human       &  \\ \cmidrule{3-9} 
            &            & Metaphor     & MABL \cite{kabra-etal-2023-multi}   & ID & WA   & Y    &             -&  \\ \cmidrule{2-9} 
            & NLR        & NLI           & XNLI \cite{conneau_xnli_2018}               & FIL, TH, VI & WA   & N      & Human       &              \\
            &            &               & IndoNLI \cite{mahendra_indonli_2021}            & ID & WA & Y      &             -&              \\
            &            &               & IndicXNLI \cite{aggarwal-etal-2022-indicxnli}          & TA      & WA   & N      & Machine     &              \\ \cmidrule{3-9} 
            &            & Causal        & Balanced COPA \cite{kavumba-etal-2019-choosing} & FIL   & WA     & N    & Human         &              \\
            &            &               & XCOPA \cite{ponti_xcopa_2020}              & ID, TA, TH, VI & WA & N      & Human       &            \\ \cmidrule{2-9} 
            & NLG        & Summarisation & XL-Sum \cite{hasan_xl-sum_2021}             & FIL, ID, TA, TH, VI  & Rouge-L             & Y      &             -&              \\ \cmidrule{3-9} 
            &            & Translation   & FLORES \cite{nllb2022}             & FIL, ID, TA, TH, VI & MetricX-wmt24       & N      & Human       &    
            \\
            \midrule
LLM-specifics & Instruction following & SEA-IFEval & SEA-IFEval & FIL, ID, TH, TA, VI  & LNA & N & Human & \checkmark \\ \cmidrule{2-9} 
             & Chat capability & SEA-MTBench & SEA-MTBench & FIL, ID, TA, VI & WR & N & Human & \checkmark \\
             &                 &             & MT-Bench Thai \cite{MTBench_Thai} & TH & WR & N & Human &        \\
\midrule
\multirow[t]{2}{*}{SEA Linguistics}    & Linguistic   & Pragmatics & LINDSEA               & ID, TA & WA & Y      &             -& \checkmark    \\
           & Diagnostics      &  Syntax & LINDSEA               & ID, TA & WA & Y      &             -& \checkmark   \\ \hline
SEA Culture & Cultural representation & KALAHI & KALAHI \cite{montalan2024kalahihandcraftedgrassrootscultural} & FIL & WA & Y & -& \\
\midrule
Safety      & Toxicity   & Toxicity Detection & MLHSD \cite{ibrohim_multi-label_2019}              & ID & WA & Y      &             -&              \\
            &            &               & Thai Toxicity Tweet \cite{sirihattasak2018annotation} & TH & WA   & Y      &             -&              \\
            &            &               & ViHSD \cite{luu2021large}              & VI & WA   & Y      &             -&              \\
            &            &               & PH Elections Toxicity \cite{cabasag2019hate}       & FIL   & WA                 & Y    &          -&              \\ \bottomrule
\end{tabular}%
}
\caption{List of datasets used in SEA-HELM, which are either created from scratch, translated or collated from other sources. We curate datasets that are created in the native language (indicated with "-") and translated datasets (either by humans or machines). Metrics: WA (weighted accuracy), LNA (language-normalised accuracy), WR (Win rate against \textit{gpt-3.5-turbo-0125} as judged by \textit{gpt-4-1106-preview}), Rouge-L (multilingual ROUGE implementation from XL-Sum \cite{hasan-etal-2021-xl}), MetricX-wmt24 ( \textit{metricx-24-hybrid-xxl-v2p6-bfloat16} model \cite{juraska-etal-2024-metricx}).}
\label{tab:dataset}
\end{table*}

\subsection{Core pillars} \label{competencies}
\subsubsection{NLP classics} \label{classical}

First, for the Natural Language Understanding (NLU) competency, we include QA (extractive question answering) and sentiment analysis tasks. Second, for the Natural Language Generation (NLG) competency, we include translation (English to native language and native language to English) and abstractive summarisation tasks. Third, for the Natural Language Reasoning (NLR) competency, we include causal reasoning and natural language inference (NLI) tasks.\footnote{It should be noted that reasoning as a competency is broad and often a reasoning task requires many specific types of reasoning skills \citep{espejel2023gpt35,huang_towards_2023,qiao_reasoning_2023, xu_are_2023, yu_natural_2023}.}

We selected datasets that comprised of data originally written in the native language as far as possible. Otherwise, existing datasets in English were carefully translated by native speakers.\footnote{Annotator demographics are not included in this paper for anonymity, demographics can be provided upon request.} This is important because translated datasets often contain elements of translationese \citep{gellerstam1986translationese}, which can differ significantly from natively written text \citep{baker_corpus_1993, lembersky_language_2012, volansky_features_2015, riley_translationese_2020}. 

\subsubsection{LLM-specifics} 
With LLMs enabling unprecedented NLP applications, there is a need to develop automated, dedicated evaluation metrics for these higher-order tasks. SEA-HELM focuses on two specific capabilities - the ability to follow human instructions specifying a particular format expected in the given responses, and the ability to hold human-like conversations. The former can be evaluated using simpler rule-based checkers that examine the format of the LLM's responses, while the latter requires us to model subjective human preferences, and thus employs the LLM-as-a-judge paradigm \cite{zheng2023judging, kim2024prometheusinducingfinegrainedevaluation}. 

\textbf{SEA-IFEval} is an instruction-following benchmark we created collaboratively with native speakers. It was manually translated from the English IF-Eval benchmark \citep{zhou2023instruction} and, crucially, localised to fit the linguistic and cultural nuances of SEA languages. Manual translations ensured faithful and accurate linguistic representation, while localisation ensured cultural authenticity and removed any unintended or inherent biases. Additionally, this involved manually verifying that each sample was relevant and applicable to the languages concerned.

Specifically, to create the SEA-IFEval dataset, we first filtered out instructions that could not reasonably apply to most SEA languages. For example, prompts asking to change the capitalisation or punctuation do not make sense in many scripts in the region, such as Burmese, Tamil, or Thai. We also changed instructions that required inclusion of a stated quantity of instances of letters to instead require a stated quantity of instances of numbers as it was not easy to localise them for non-Latin scripts. An example of this change is to convert the instruction of “[respond] with the letter ‘l’ appearing at least 6 times” to “[respond] with the number ‘4’ appearing at least 6 times. Thus, by filtering out and adapting instructions given the SEA context, we ensure a fair basis of comparison for the instruction-following competency. The final categories included in the SEA-IFEval dataset are listed in Table \ref{tab:sea_if_eval_subcategories} in Appendix \ref{sec:llm_specifics_pillar}.

The accuracy with which LLMs follow the exact instruction following requirements is calculated using the same verifiers as those found in the English IF-Eval benchmark \cite{zhou2023instruction}. The model's accuracy is then adjusted to penalise instructions that responded in the wrong language by multiplying its accuracy with the rate at which it responds in the correct target language.

\textbf{SEA-MTBench} is a manually translated and localised version of the popular MT-Bench dataset \cite{zheng2023judging}, which also introduced the paradigm of LLM-as-a-Judge to approximate human preferences. We chose the reference-guided grading approach, where we compare the win-rate of each candidate model against a fixed reference model, namely \textit{gpt-3.5-turbo-0125}. The models' responses were compared with the reference response using \textit{gpt-4-1106-preview} as the judge model. This setup sees the number of judge calls scale linearly with the number of models being compared, whereas pairwise comparisons would scale quadratically.

Models receive an initial prompt based on a category such as creative writing, mathematics, STEM or humanities. They are then given a follow-up instruction which related to the initial prompt, and are expected to respond appropriately. Finally, a model’s responses to both the initial and follow-up prompts are evaluated as a whole given accuracy, relevance, and coherence as judging criteria. Results are reported based on each model's average win rate against the reference model.

\subsubsection{SEA linguistics}
As one of the five core evaluation pillars, \textbf{LINDSEA }(\textbf{LIN}guistic \textbf{D}iagnostics for \textbf{S}outh\textbf{E}ast \textbf{A}sian languages) is a high quality, manually-crafted linguistic dataset that systematically diagnoses models' language proficiency and grammatical understanding based on a granular taxonomy of syntactic, semantic and pragmatic phenomena. It is also the first to be created for SEA languages. LINDSEA provides fine-grained evaluation of a model's linguistic abilities, akin to the diagnostic dataset of GLUE \citep{wang_glue_2018} and BLiMP \cite{warstadt_blimp_2020}, the linguistic diagnostic dataset for HELM.

The design of LINDSEA is based on three principles: \textbf{breadth}, \textbf{depth}, and \textbf{quality}. Given the increasingly complex tasks that LLMs are expected to perform and the importance of both natural language input and output in users' interactions with LLMs, it is crucial that we are able to comprehensively evaluate and quantify models' understanding of the myriad aspects of language. To that end, LINDSEA is designed to cover a wide gamut of linguistic phenomena (\textbf{breadth}). In designing LINDSEA to have sufficient linguistic coverage, we also conducted an extensive survey of the literature on linguistic phenomena in our target languages and used our findings to taxonomise each linguistic phenomenon to have multiple categories and subcategories for more fine-grained analyses (\textbf{depth}). That is, rather than using a small set of lexical items and grammatical rules to automatically generate large numbers of test sentences, the examples in LINDSEA are handcrafted from scratch by linguists in collaboration with native speakers and reviewed iteratively to ensure that they sound natural, are semantically coherent and target the relevant phenomena effectively (\textbf{quality}). 

While there are existing syntactic and semantic diagnostic datasets for English \citep{warstadt_blimp_2020,jeretic_are_2020,liu_adjective_2023}, Mandarin \citep{xiang_climp_2021} and even Japanese \citep{someya-oseki-2023-jblimp}, none yet exist for SEA languages, and, to our knowledge, there has yet to be such an extensive coverage of linguistic phenomena in any dataset. More details about the individual subcategories and literature reviewed can be found in Appendix \ref{tab:lindsea_references}.

\subsubsection{SEA culture}
Cultural representation and bias have also become increasingly important with LLM use \citep{adilazuarda-etal-2024-towards} since a lack thereof can potentially cause social harm \citep{solaiman_evaluating_2023}. The gravity of the risks involved has prompted multiple studies in this area \citep{naous_having_2023,ramesh-etal-2023-fairness,ramezani_knowledge_2023}. 

Much prior work on analysing or evaluating cultural representation in LLMs is created in ways that do not start from consulting the communities of speakers of the language but instead starts from reference sources that aggregate cultural knowledge, opinions, and values at a population level. This ``top-down'' approach also demonstrates a notable lack of emphasis on the perspectives, decisions, and actions that individuals take as they participate within their communities or navigate their daily lives. For example, \citet{durmus_towards_2023} frame cultural representation as whether models exhibit values aligned with those of people from different countries, as extracted from large-scale surveys, such as the World Values Survey\footnote{\url{https://www.worldvaluessurvey.org/wvs.jsp}} and the Pew Global Attitudes Survey.\footnote{\url{https://www.pewresearch.org/expertise/international-attitudes/}} Such top-down value extraction can involve subject matters that are primarily determined by people who are not members of the community, and thus may not be fully representative of the community's concerns and lived experiences. 

There are also top-down SEA efforts, such as SeaEval \citep{wang2023seaeval} as well as SeaExam and SeaBench \citep{liu2025seaexamseabenchbenchmarkingllms}, that primarily serve as tests for factuality and general local knowledge. Such datasets are inherently limited and cannot comprehensively represent the complex nature of culture \cite{causadias2020culture}, even if they can capture aggregated value alignment. Thus, we place emphasis on a strong participatory approach that includes the native speaker communities in order to authentically represent the target culture.

\citet{hershcovich_challenges_2022} suggest that a culture can be defined by its shared cultural common ground or a shared body of knowledge within the community, while \citet{swidler1986culture} proposes that culture is expressed in the strategies of action or ``toolkit'' that people use to navigate their personal and social lives. Thus, to probe models for their understanding of cultural knowledge and to evaluate if models can appropriately apply cultural knowledge or values, we have included \textsc{Kalahi} \cite{montalan2024kalahihandcraftedgrassrootscultural}, which was developed using a participatory approach, under the SEA Culture pillar. The \textsc{Kalahi} dataset is designed to determine if LLMs can provide culturally-relevant responses to culturally-specific situations that Filipino people can reasonably encounter in their daily lives, and the dataset is composed of 150 high-quality Filipino-language\footnote{Filipino is the national language of the Philippines \cite{philippines1987} and is primarily based on Standard Tagalog \cite{KWF1996}} prompts created in collaboration with native Filipino speakers\footnote{For further details on \textsc{Kalahi}, please refer to \citep{montalan2024kalahihandcraftedgrassrootscultural}} (see Table \ref{tab:example_cultural_appropriateness} in Appendix \ref{sec:sea_culture_pillar} for an example of its implementation in SEA-HELM). 

\subsubsection{Safety}
Multilingual inputs, especially when given in lower-resource languages such as those in the SEA region, can increase the likelihood that LLMs produce unsafe responses \citep{song2024multilingual,shen2024language}. Thus, tailoring safety benchmarks that cater to SEA languages and cultures is critical in ensuring that users who interact with models in those languages are protected from harmful and unsafe outputs (e.g. hate speech). In this regard, we also aim to foster and enhance representativeness and inclusivity by curating datasets that are relevant to SEA languages. After a thorough survey of the available datasets similar to the one performed in section \ref{classical}, we decided to include toxicity detection as the first task (of many planned for the future) under the \textsc{Safety} pillar.\footnote{For further details on the toxicity dataset, please refer to \citep{leong2023bhasa}.} Currently, we cover Indonesian, Thai, Vietnamese, and Filipino for this task. This serves as the starting point that will lead to more complete and comprehensive SEA safety benchmarks in SEA-HELM.

\subsection{SEA-HELM leaderboard} 
Given the large number of tasks in SEA-HELM, it can be challenging to determine the overall performance of a given model. There is a need to aggregate task scores. We believe that the aggregated scores should be presented in a clear and transparent manner that allows users and developers to delve deeper into each aggregation in a maximally informative manner as well. To facilitate this, we have released the SEA-HELM leaderboard as part of the SEA-HELM suite. The leaderboard can be accessed at \url{https://leaderboard.sea-lion.ai/}.

The leaderboard presents the results in three separate views -- an overall view containing the SEA average as well as the overall language scores, a language view showing the average competency scores for that language, and a detailed view containing the normalised scores for each task (Instruction-tuned models:  Table \ref{tab:instruction_tuned_model_performance}, Base models: Tables \ref{tab:small_base_model_performance} and \ref{tab:large_base_model_performance}). Additionally, it includes the results from both the pre-trained and the instruction-tuned variants of each model across a wide range of model sizes. See Appendix \ref{appendix_leaderboard} for more examples of results visualisations in the leaderboard.

\begin{table*}[t]
\centering
\resizebox{\textwidth}{!}{%
\begin{tabular}{lll|cccccc}
\toprule
Model & Size & Supported SEA Languages & SEA Average & ID & VI & TH & TA & FIL \\
\midrule
gpt-4o-2024-08-06 & - & - & 68.9 & 74.5 & 68.1 & 64.7 & 64.2 & 73.0 \\
DeepSeek-R1 & 671B & - & 68.3 & 72.1 & 68.0 & 62.3 & 66.5 & 72.6 \\
\midrule
llama3.1-70b-cpt-sea-lionv3-instruct & 70B & TH,VI,TL,TA,ID & 67.1 & 71.1 & 68.3 & 61.9 & 63.9 & 70.5 \\
gemma-2-27b-it & 27B & - & 65.4 & 67.9 & 61.6 & 63.1 & 64.0 & 70.5 \\
Llama-3.3-70B-Instruct & 70B & TH & 64.9 & 70.4 & 68.3 & 59.7 & 56.6 & 69.4 \\
Qwen2.5-72B-Instruct & 72B & TL,TH,VI,ID & 62.1 & 69.8 & 65.8 & 63.5 & 44.9 & 66.6 \\
Llama-3.1-70B-Instruct & 70B & TH & 61.9 & 67.8 & 63.4 & 56.1 & 53.1 & 69.1 \\
Qwen2.5-32B-Instruct & 32B & TL,TH,VI,ID & 61.5 & 69.0 & 65.7 & 61.1 & 49.0 & 62.8 \\
Meta-Llama-3-70B-Instruct & 70B & - & 54.5 & 57.4 & 53.5 & 50.4 & 52.2 & 58.8 \\
aya-expanse-32b & 32B & VI,ID & 50.2 & 68.1 & 62.0 & 40.4 & 26.9 & 53.6 \\
Sailor2-20B-Chat & 20B & TL,TH,VI,ID & 39.2 & 55.8 & 41.1 & 39.6 & 14.3 & 45.5 \\
\midrule
gemma2-9b-cpt-sea-lionv3-instruct & 9B & TH,VI,TL,TA,ID & 63.2 & 65.1 & 64.2 & 59.9 & 59.6 & 67.2 \\
gemma-2-9b-it & 9B & - & 60.0 & 62.7 & 60.3 & 57.4 & 57.6 & 62.2 \\
llama3.1-8b-cpt-sea-lionv3-instruct & 8B & TH,VI,TL,TA,ID & 55.7 & 60.6 & 59.9 & 53.7 & 47.7 & 56.5 \\
Qwen2.5-7B-Instruct & 7B & TL,TH,VI,ID & 46.1 & 61.1 & 56.6 & 55.2 & 14.7 & 43.0 \\
Sailor2-8B-Chat & 8B & TL,TH,VI,ID & 42.2 & 47.9 & 44.4 & 39.4 & 27.0 & 52.4 \\
Llama-3.1-8B-Instruct & 8B & TH & 39.7 & 53.4 & 49.8 & 39.3 & 13.4 & 42.6 \\
SeaLLMs-v3-7B-Chat & 7B & TH,VI,TL,TA,ID & 39.6 & 47.9 & 51.8 & 47.6 & 9.5 & 41.4 \\
Meta-Llama-3-8B-Instruct & 8B & - & 34.9 & 46.7 & 46.4 & 35.8 & 12.4 & 33.3 \\
aya-expanse-8b & 8B & VI,ID & 33.4 & 59.0 & 56.4 & 20.2 & 10.0 & 21.5 \\
\bottomrule
\end{tabular}
}
\caption{SEA-HELM normalised scores of instruction tuned models, separated into larger (>10 B parameters) and smaller (<10 B parameters) models. Results for \textit{gpt-4o-2024-08-06} and \textit{DeepSeek-R1} are provided as reference. Supported SEA Languages comprises of tested languages that are reported on the respective model cards or reports. }
\label{tab:instruction_tuned_model_performance}
\end{table*}

\subsection{SEA-HELM evaluation design choices}
\subsubsection{Task prompt format}
For each task, we opted to be explicit in instructing the LLM to output the answer following a specified format. This is done by prompting the model to return its response with an answer tag. Table \ref{tab:example_qa} shows an example of such a prompt and the expected generation from the model (see Appendices \ref{sec:nlp_classics_pillar} to \ref{sec:safety_pillar} for all prompts used in SEA-HELM), and note that the prompt explicitly states that the answer tag \textit{Jawaban} (which is the Indonesian word for `answer') must be prefixed to its answer (see Appendix \ref{sec:nlp_classics_pillar}). The answer can then be extracted using regular expressions and compared against the corresponding gold label. Crucially, if the model fails to append the answer tag to its answer, the model is deemed to have given a null response. This approach allows for more efficient automatic evaluation at scale, even for models that may tend to have lengthy outputs. 

It should also be noted that each prompt is given in the target language rather than in English (except for the English tasks) and is manually translated by native speakers of each language. In our view, for full SEA languages support, an LLM should be able to output coherent responses and should also be able to correctly interpret native instructions. Instruction-tuned models were evaluated in a zero-shot manner while pre-trained models are evaluated with 5-shot examples.

\subsubsection{Normalisation of metrics}

As per equation \ref{eq:score_normalisation}, the normalised metric score ($s_{norm}$) is calculated by subtracting the random baseline score ($s_{baseline}$) from the raw metric score ($s_{raw}$). This is then divided by the difference between the maximum score ($s_{max}$) and the random baseline score. The range of the normalised score is then scaled to a range of 0--100 \citep{cohere2025towards}. For negative normalised score, we set the score to 0.

For multiple-choice task with $n$ options, the baseline score using a random strategy is $(100/n)\%$. For open ended questions with a score range between 0--1, we set the guessing score to 0\% and the best possible score to 100\%.

\begin{equation}
    s_{norm}=\max(100*\frac{s_{raw}-s_{baseline}}{s_{max}-s_{baseline}}, 0)
\label{eq:score_normalisation}
\end{equation}

\subsubsection{Aggregation of metrics}
First, metrics are aggregated at the competency level. The normalised scores of each tasks belonging to each competency are averaged to calculate the competency score (The list of tasks in each competency can be found in Table \ref{tab:dataset}). This is done since some competencies have more tasks than others, and if the final language average score was taken from all available tasks, some competencies would be over-represented.

Next, the language score is calculated by taking the average of the competency scores for each language. This provides an indication of how well the models are performing at a per-language basis

Finally, a SEA average score is calculated as the mean of all the language score in SEA-HELM.

\subsection{Evaluation of LLMs with SEA languages support}
\begin{figure*}[t]
    \begin{subfigure}[c]{\textwidth} 
        \includegraphics[width=\textwidth]{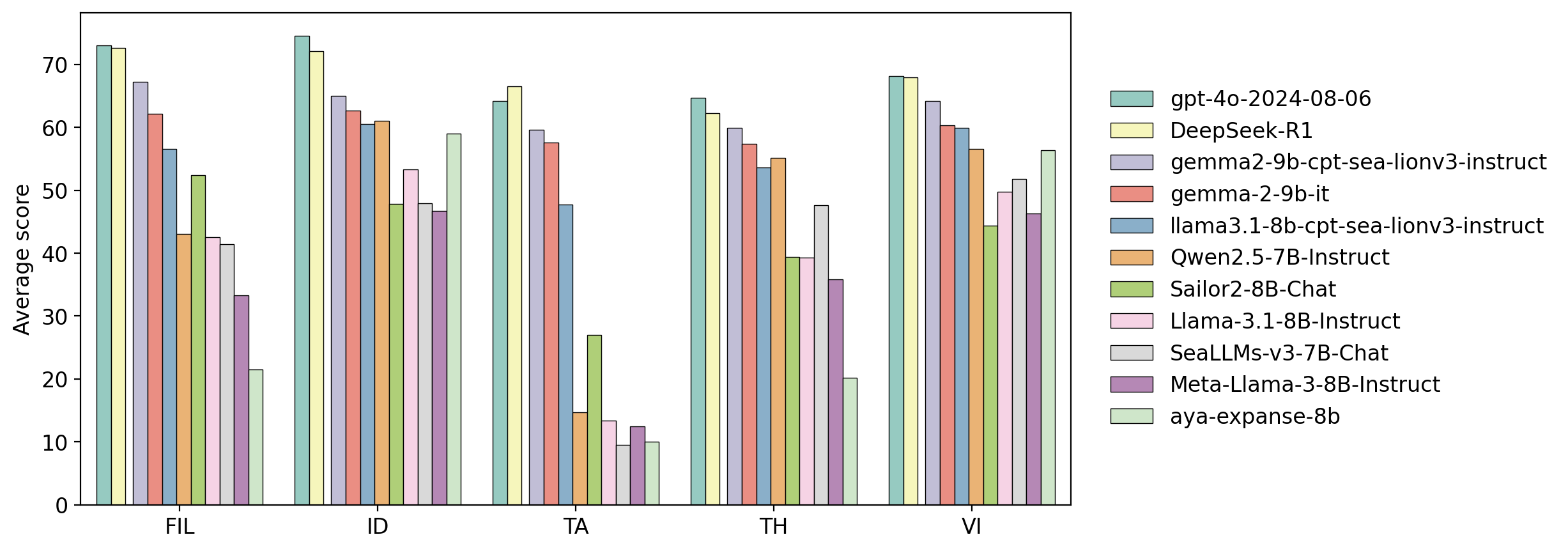}
        \caption{Average SEA-HELM scores by language.}
        \label{fig:leaderboard_graph}
    \end{subfigure}
    \hfill
    \begin{subfigure}[c]{0.38\textwidth}
        \includegraphics[width=\textwidth]{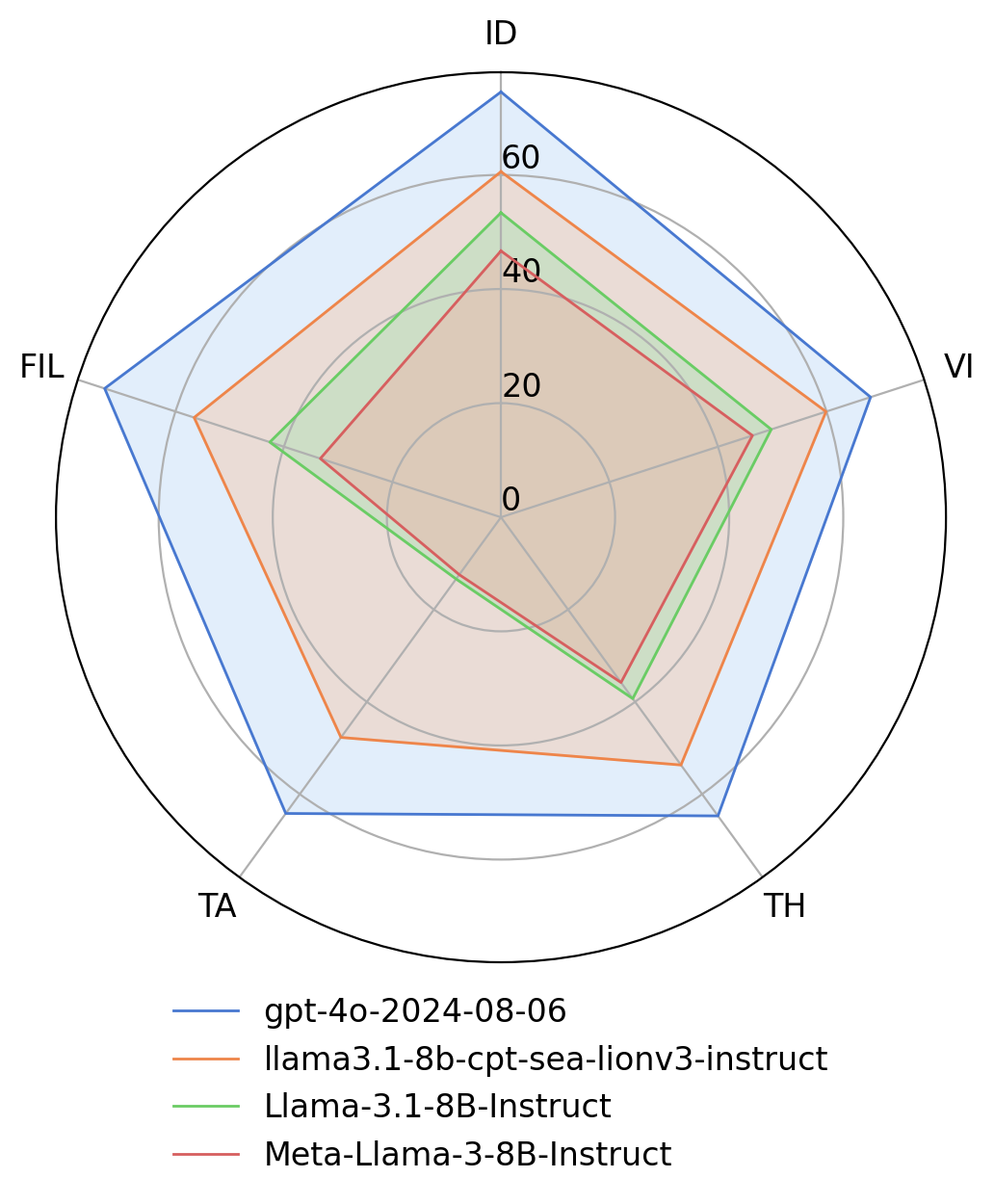}
        \caption{Improvements in the LLaMA family of models}
        \label{fig:llama_spiderplot}
    \end{subfigure}
    \hfill
    \begin{subfigure}[c]{0.6\textwidth}
        \includegraphics[width=\textwidth]{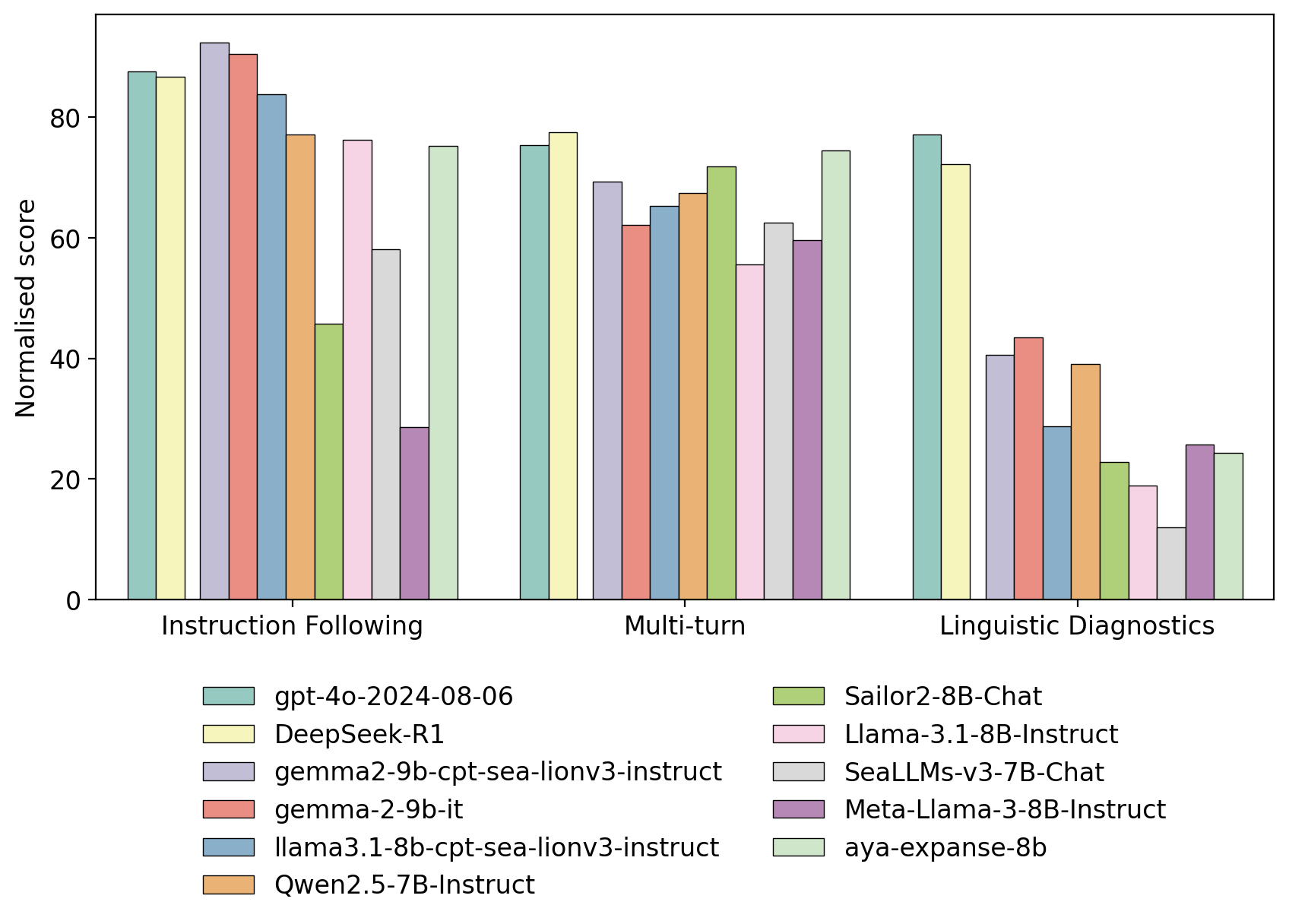}
        \caption{Indonesian performance on Instruction Following, Multi-turn, and Linguistic Diagnostics}
        \label{fig:ifeval_mtbench_lindsea}
    \end{subfigure}
    \caption{Models sizes range between 7-9B parameters. Results for \textit{gpt-4o-2024-08-06} and \textit{DeepSeek-R1} are provided as reference.}
\end{figure*}

We used SEA-HELM to perform a comprehensive evaluation of several models that have some level of proficiency in the SEA languages (Figure \ref{fig:leaderboard_graph}). Examples of such models include the Sailor family \citep{sailor2report}, SeaLLMs family \citep{zhang2024seallms} and the SEA-LION family of models.\footnote{\url{https://huggingface.co/collections/aisingapore/sea-lionv3-672589a39cdadd6a5b199581}} Of all the open-sourced models tested, \textit{DeepSeek-R1} was the strongest performing model and showed comparable results to \textit{gpt-4o-2024-08-06} on the SEA-HELM suite (Figure \ref{fig:leaderboard_graph}, Table \ref{tab:instruction_tuned_model_performance}). Given the large size of \textit{DeepSeek-R1} (671B), finding comparisons to other models that support SEA languages was not trivial. Thus, we chose to focus our discussion on LLMs of sizes ranging between 7-9B parameters (See Table \ref{tab:instruction_tuned_model_performance} for the full list of models) and used both the \textit{DeepSeek-R1} and \textit{gpt-4o-2024-08-06} as the reference models for what was the best available open-sourced and closed-sourced model respectively.

We also observe that the gap between the smaller models and the reference models is closing, as evidenced by the smaller gap between \textit{gpt-4o-2024-08-06} with the newer LLaMA model \textit{Llama-3.1-8B-Instruct} as compared to the earlier LLaMA model \textit{Meta-Llama-3-8B-Instruct} (Figure \ref{fig:leaderboard_graph}, \ref{fig:llama_spiderplot}, Table \ref{tab:instruction_tuned_model_performance}). Notably, continued pre-training and further tuning for the SEA languages resulted in this gap closing even further (Figure \ref{fig:llama_spiderplot}; \textit{llama3.1-8b-cpt-sea-lionv3-instruct} compared against \textit{Llama-3.1-8B-Instruct}). This suggests that there is still room for improvement and spending the effort to train LLMs to target the specific SEA languages can result in models that are more suited for the SEA region and thus work better for the region's use cases.

The choice of model family is also important. The Gemma2 family of models (\textit{gemma-2-9b-it}) and the continued pre-trained/fine-tuned model (\textit{gemma2-9b-cpt-sea-lionv3-instruct}) performed the best of all the evaluated models. One possible explanation could be the choice of tokenizer (256k vocabulary size) which was shown to be associated with better downstream performance for most languages, especially in multilingual settings \citep{ali2024tokenizerchoicellmtraining}.

Our SEA-HELM suite reveals that the performance of the LLMs for Tamil and Filipino were relatively poor (Figure \ref{fig:leaderboard_graph}, Table \ref{tab:instruction_tuned_model_performance}). The exceptions were the Gemma2-based models and \textit{llama3.1-8b-cpt-sea-lionv3-instruct}. This was likely due to the lack of support for these languages in many of these models and indicates the limited capabilities of LLMs in these lower-resource languages. As mentioned previously, supporting these languages through additional training can improve the capabilities of LLMs in these languages.

Additionally, Figure \ref{fig:ifeval_mtbench_lindsea} illustrates why it is necessary to adopt a holistic approach that evaluates models' competencies under the five pillars. As a first example, note that \textit{Meta-Llama-3-8B-Instruct}'s performance for SEA-IFEval was much lower (relatively speaking) as compared to LINDSEA and SEA-MTBench. This indicated that while the model exhibited linguistic understanding and has some LLM-specific capabilities, it was still lacking in native instruction-following competency. As another example, \textit{Sailor2-8B-Chat} was observed to perform exceptionally well for SEA-MTBench, which was indicative of more coherent and relevant native multi-turn competency, but had much poorer performance for LINDSEA and SEA-IFEval, which indicates that it was much weaker in terms of linguistic understanding as well as native instruction following.

Full results can be viewed on our leaderboard, and readers can also reference Appendix \ref{appendix_leaderboard}.

\section{Conclusion}
In conclusion, we introduce SEA-HELM, a holistic evaluation suite that caters to the SEA languages and cultures. To achieve a comprehensive evaluation suite, SEA-HELM was designed around the following five core pillars: (1) \textsc{NLP Classics}, (2) \textsc{LLM-specifics}, (3) \textsc{SEA Linguistics}, (4) \textsc{SEA Culture}, (5) \textsc{Safety}.

Additionally, the SEA-HELM leaderboard is intended to serve as a comprehensive and regularly maintained resource for AI researchers in academia and industry. Our results show that there are still significant gaps between high-resource languages such as English and the mid- to low-resource languages in Southeast Asia, including several with official status and millions of speakers. However, these results also show that the dedicated fine-tuning of LLMs with parameter sizes between 7 and 9 billion can significantly narrow the gap with respect to the much larger state-of-the-art models such as \textit{GPT-4o} and \textit{DeepSeek-R1}. Thus, by calling attention to the realistic, localised needs of SEA languages, we encourage more concentrated efforts on data collection, curation and fine-tuning of dedicated lighter-weight LLM solutions for the region. 

\section{Future work}
We recognise that although SEA-HELM currently covers Filipino, Indonesian, Tamil, Thai, and Vietnamese, we still have much more work to do and therefore we are committed to iteratively expanding each pillar.

Specifically, we plan to expand SEA-HELM to include a broader range of languages, cultures, tasks and models to encourage stronger SEA representation in models. We also seek to explore additional factors such as automatic LLM evaluations. This will enable an even more comprehensive evaluation of LLM performance across a wider variety of contexts for SEA languages, especially for low-resource languages in the region such as Burmese, Khmer and Lao.

\section*{Limitations} 
While aiming to achieve holistic and authentic evaluations for LLMs in Southeast Asia, SEA-HELM is not yet exhaustive in its coverage of languages and tasks. As we iteratively improve the coverage of the various SEA-HELM pillars, more language and cultural coverage will added and refined in the future.

Our leaderboard results are based on single model runs. However, as we have assumed deterministic model behaviour and set every model's temperature to 0, we did not publish error bars for the results, in line with most other benchmarks in the field.

Under the safety pillar of SEA-HELM, we also acknowledge that passing our evaluations with a high score is not necessarily a guarantee of the safety of an LLM in the SEA context, as it is not possible to enumerate every type of unsafe response in real-world scenarios. Thus, passing the safety evaluations in SEA-HELM must be seen as a necessary but not sufficient requirement for ensuring safety in real-world LLM applications.

\section*{Ethics Statement}

SEA-HELM is grateful to our Quality Assurance (QA) team, consisting of paid native speakers of SEA languages who worked as translators and annotators, enabling us to construct localised datasets needed for this research. 
The SEA-HELM project has received full and official approval from our university's internal review board after undergoing a rigorous review process, and the compensation and working hours for all members involved were established in full compliance with the prevailing university guidelines and applicable regulations in the country where this research is conducted.

Our QA team was recruited through public advertisements that clearly outlined the estimated workload and hourly pay, consistent with all relevant legal and regulatory requirements. The team is composed predominantly of students at local universities and members of the public, all of whom are adults who satisfy the legal age requirements for employment, as defined by the labour laws of the country. Although participants are compensated for their participation, their involvement in the study is entirely voluntary. Any personally identifiable information (PII) is removed from the data collected and will not be tied to their identity.

We did not estimate that their work involved significant risks of exposure to offensive material, as they were not involved in the construction of sensitive data such as those under the \textsc{Safety} pillar. Nonetheless, they were encouraged to report inappropriate samples and were given the option to withdraw their participation at any time, including after its completion, without any negative consequences or loss of benefits. If they chose to withdraw, they could do so without providing any reason, and all data collected from the withdrawn individual were excluded from this research.

We do not foresee negative social impacts from our research, for our research introduces evaluation datasets in SEA languages by working with native speakers, paying due respect to local cultural sensitivities. We thus do not believe that our research will contribute to over-generalisations regarding SEA cultures.

\section*{Acknowledgments}
This research/project is supported by the National Research Foundation, Singapore under its National Large Language Models Funding Initiative. Any opinions, findings and conclusions or recommendations expressed in this material are those of the author(s) and do not reflect the views of National Research Foundation, Singapore.

We would like to extend our heartfelt thanks to our colleagues at AI Singapore, especially AI Products and SEA-LION team, for their contributions in enriching the discussions and for generously sharing their valuable and insightful feedback. In the same way, we would like to thank the HELM team for their unwavering support of our initiatives from the very beginning.

We would like to also express our sincere appreciation to the native speakers of SEA languages involved in this project for their steadfast commitment and the generous time they devoted to assisting us with human evaluation, annotation, and translation.

\bibliography{references}
\bibliographystyle{acl_natbib}

\appendix
\onecolumn

\setcounter{table}{0}
\renewcommand*\thetable{\Alph{section}.\arabic{table}}

\newpage
\section{Normalised SEA-HELM scores}
\label{appendix_leaderboard}

\begin{table}[htp]
\centering
\resizebox{\textwidth}{!}{%
\begin{tabular}{lll|cccccc}
\toprule
Model & Size & Supported SEA Languages & SEA Average & ID & VI & TH & TA & FIL \\
\midrule
gemma2-9b-cpt-sea-lionv3-base & 9B & TH,VI,TL,TA,ID & 55.2 & 59.5 & 52.8 & 51.7 & 65.2 & 46.7 \\
gemma-2-9b & 9B & - & 49.1 & 51.9 & 47.5 & 47.3 & 57.9 & 40.9 \\
Sailor2-8B & 8B & TL,TH,VI,ID & 49.0 & 55.5 & 49.6 & 51.8 & 48.7 & 39.6 \\
Qwen2.5-7B & 7B & TL,TH,VI,ID & 45.5 & 51.3 & 54.2 & 50.1 & 37.8 & 33.9 \\
llama3.1-8b-cpt-sea-lionv3-base & 8B & TH,VI,TL,TA,ID & 45.2 & 49.2 & 48.3 & 46.6 & 46.2 & 35.6 \\
SeaLLMs-v3-7B & 7B & TH,VI,TL,TA,ID & 41.5 & 47.0 & 50.4 & 47.6 & 28.6 & 34.0 \\
Meta-Llama-3.1-8B & 8B & TH & 40.3 & 45.9 & 42.8 & 39.0 & 42.4 & 31.6 \\
Meta-Llama-3-8B & 8B & - & 37.4 & 45.1 & 41.4 & 41.0 & 30.5 & 28.9 \\
\bottomrule
\end{tabular}
}
\caption{SEA-HELM normalised scores of base models with sizes between 7-9B parameters. Supported SEA Languages comprises of tested languages that are reported on the respective model cards or reports.}
\label{tab:small_base_model_performance}
\end{table}

\begin{table}[htp]
\centering
\resizebox{\textwidth}{!}{%
\begin{tabular}{lll|cccccc}
\toprule
Model & Size & Supported SEA Languages & SEA Average & ID & VI & TH & TA & FIL \\
\midrule
llama3.1-70b-cpt-sea-lionv3-base & 70B & TH,VI,TL,TA,ID & 58.7 & 61.9 & 59.0 & 55.4 & 66.3 & 50.6 \\
Qwen2.5-72B & 72B & TL,TH,VI,ID & 58.6 & 64.5 & 61.2 & 57.2 & 59.5 & 50.4 \\
Qwen2.5-32B & 32B & TL,TH,VI,ID & 55.7 & 66.7 & 58.9 & 58.2 & 53.3 & 41.4 \\
Meta-Llama-3.1-70B & 70B & TH & 55.5 & 61.2 & 54.3 & 52.4 & 63.3 & 46.1 \\
gemma-2-27b & 27B & - & 53.9 & 56.8 & 52.1 & 53.6 & 61.7 & 45.1 \\
Sailor2-20B & 20B & TL,TH,VI,ID & 51.3 & 61.9 & 55.1 & 52.2 & 47.5 & 39.6 \\
Meta-Llama-3-70B & 70B & - & 50.9 & 59.3 & 52.1 & 49.7 & 47.4 & 45.8 \\
\bottomrule
\end{tabular}
}
\caption{SEA-HELM normalised scores of base models with sizes between 20-72B parameters. Supported SEA Languages comprises of tested languages that are reported on the respective model cards or reports.}
\label{tab:large_base_model_performance}
\end{table}

\begin{table}[htp]
\centering
\resizebox{\textwidth}{!}{%
\begin{tabular}{lrrrr|r}
\toprule
Model Group & \small Size Range & \small No. models & \small GPU hours & \small No. H100 GPUs  & \small Total H200 GPU hours \normalsize \\
\midrule
DeepSeek-R1 & 671B & 1 & 4 & 8 & 32 \\
Large instruction-tuned models & 20-72B & 9 & 1 & 4 & 36 \\
Large base models & 20-72B & 9 & 1 & 4 & 36 \\
Small instruction-tuned models & 7-9B & 7 & 1 & 1 & 7 \\
Small base models & 7-9B & 8 & 1 & 1 & 8 \\
\hline
Total & & & & & \textbf{84} \\
\bottomrule
\end{tabular}
}
\caption{Compute budget per run of SEA-HELM.}
\label{tab:compute_budget}
\end{table}

\newpage
\section{Examples from the NLP Classics Pillar}
\label{sec:nlp_classics_pillar}
\begin{table}[htp]
\small
\centering
\begin{tabular}{p{3.00cm} p{12cm}}
\hline
\textbf{Component} & \textbf{Text}
\\
\hline
Prompt template 
& \begin{lstlisting}
Anda akan diberikan sebuah paragraf dan sebuah pertanyaan. 
Jawablah pertanyaannya dengan mengambil jawabannya dari paragraf tersebut.

Jawablah dengan hanya menggunakan format berikut ini.
Jawaban: $ANSWER
Ganti $ANSWER dengan jawaban yang telah ditentukan.{fewshot_examples}

Paragraf:
```
{text}
```
Pertanyaan: {question}
\end{lstlisting}
\\
\hline
Text 
& Menurut para sesepuh desa, penemu desa ini adalah orang Sunda oleh karena itu desa ini diberi nama ``Bodas'' yang artinya ``putih'' dalam bahasa Sunda. Menurut sesepuh warga setempat warga setempat mempunyai pantangan dilarang berjualan nasi \& daun sirih.
\\
\hline
Question
& Dari mana asal kata Bodas?
\\
\hline
Label
& bahasa Sunda
\\
\hline
\end{tabular}
\caption{An example for Indonesian QA.}
\label{tab:example_qa}
\end{table}
\renewcommand{\arraystretch}{1.0}

\renewcommand{\arraystretch}{1.4}
\begin{table}[htp]
\small
\centering
\begin{tabular}{p{3.00cm} p{12cm}}
\hline
\textbf{Component} & \textbf{Text}
\\
\hline
Prompt template 
& 
\vspace*{1pt}
\includegraphics[width=\linewidth]{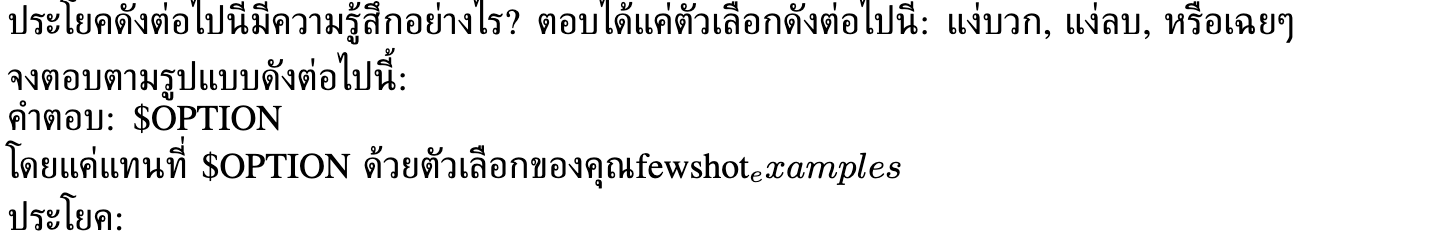}
\vspace*{1pt}
\begin{lstlisting}
```
{text}
```
\end{lstlisting}
\\
\hline
Text 
& 
\vspace*{1pt}
\includegraphics[width=\linewidth]{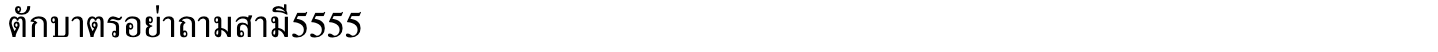}
\\
\hline
Label
& 
\vspace*{1pt}
\includegraphics[width=\linewidth]{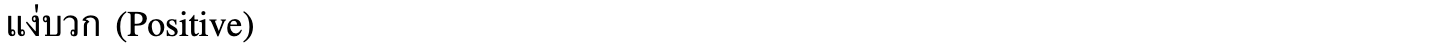}
\\
\hline
\end{tabular}
\caption{An example for Thai Sentiment.}
\label{tab:example_sentiment}
\end{table}
\renewcommand{\arraystretch}{1.0}

\renewcommand{\arraystretch}{1.4}
\begin{table}[htp]
\small
\centering
\begin{tabular}{p{3.00cm} p{12cm}}
\hline
\textbf{Component} & \textbf{Text}
\\
\hline
Prompt template 
& \begin{lstlisting}
Bibigyan ka ng dalawang pangungusap, SENTENCE_1 at SENTENCE_2. Tukuyin 
kung alin sa sumusunod na pahayag ang pinaka-angkop para sa SENTENCE_1 at 
SENTENCE_2.
A: Kung totoo ang SENTENCE_1, dapat totoo din ang SENTENCE_2.
B: Sumasalungat ang SENTENCE_1 sa SENTENCE_2.
C: Kapag totoo ang SENTENCE_1, pwedeng totoo o hindi totoo ang SENTENCE_2.

Sumagot gamit ang sumusunod na format.
Sagot: $OPTION 
Palitan ang $OPTION ng napiling sagot. 
Gumamit lang ng titik A, B, o C sa sagot mo.{fewshot_examples}

SENTENCE_1:
```
{sentence1}
```
SENTENCE_2:
```
{sentence2}
```
\end{lstlisting}
\\
\hline
Sentence 1 
& Hindi, mukhang pupunta rin ako.
\\
\hline
Sentence 2
& Mukhang hindi ako pupunta.
\\
\hline
Label
& Contradiction
\\
\hline
\end{tabular}
\caption{An example for Filipino NLI.}
\label{tab:example_nli}
\end{table}
\renewcommand{\arraystretch}{1.0}

\renewcommand{\arraystretch}{1.4}
\begin{table}[htp]
\small
\centering
\begin{tabular}{p{3.00cm} p{12cm}}
\hline
\textbf{Component} & \textbf{Text}
\\
\hline
Prompt template 
& \begin{lstlisting}
Jawablah hanya dengan menggunakan format berikut ini:
Jawaban: $OPTION
Ganti $OPTION dengan pilihan yang telah dipilih. 
Gunakan huruf A atau B saja sebagai jawabannya.{fewshot_examples}

Berdasarkan situasi yang diberikan, 
manakah dari pilihan berikut ini yang lebih mungkin 
menjadi {question_translated}?

Situasi:
```
{premise}
```
Pilihlah jawaban yang terbaik dari pilihan di bawah ini:
A: {choice1}
B: {choice2}
\end{lstlisting}
\\
\hline
Premise 
& Mata pria itu berkaca-kaca.
\\
\hline
Choice 1 
&Debu masuk ke matanya.
\\
\hline
Choice 2 
& Ia pakai kacamata.
\\
\hline
Label
& Choice 1
\\
\hline
\end{tabular}
\caption{An example for Indonesian Causal Reasoning.}
\label{tab:example_cr}
\end{table}
\renewcommand{\arraystretch}{1.0}

\renewcommand{\arraystretch}{1.4}
\begin{table}[htp]
\small
\centering
\begin{tabular}{p{3.00cm} p{12cm}}
\hline
\textbf{Component} & \textbf{Text}
\\
\hline
Prompt template 
& 
\vspace*{1pt}
\includegraphics[width=\linewidth]{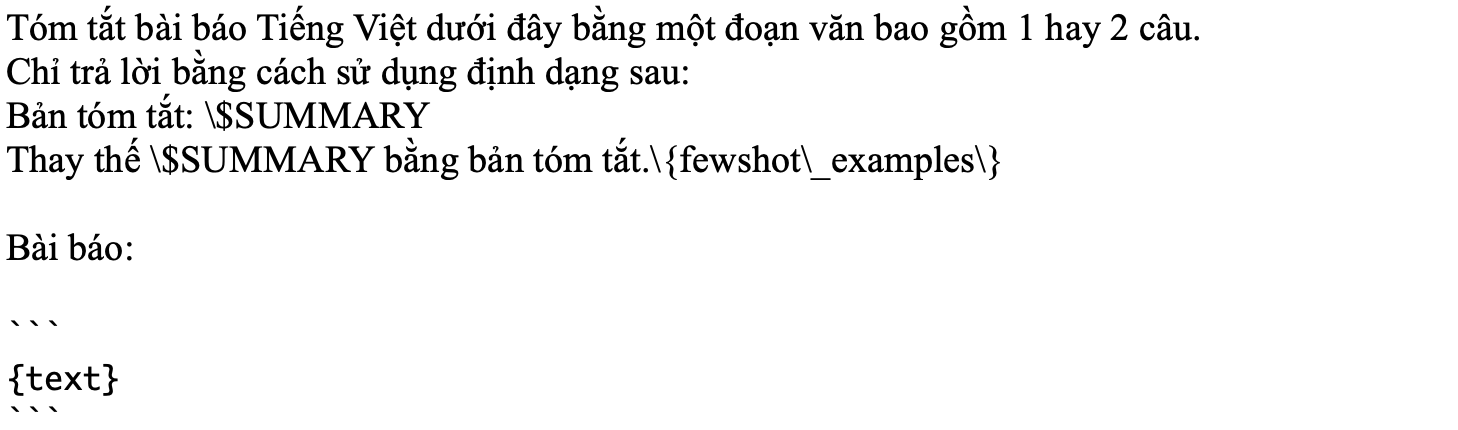}
\\
\hline
Article
&
\vspace*{1pt}
\includegraphics[width=\linewidth]{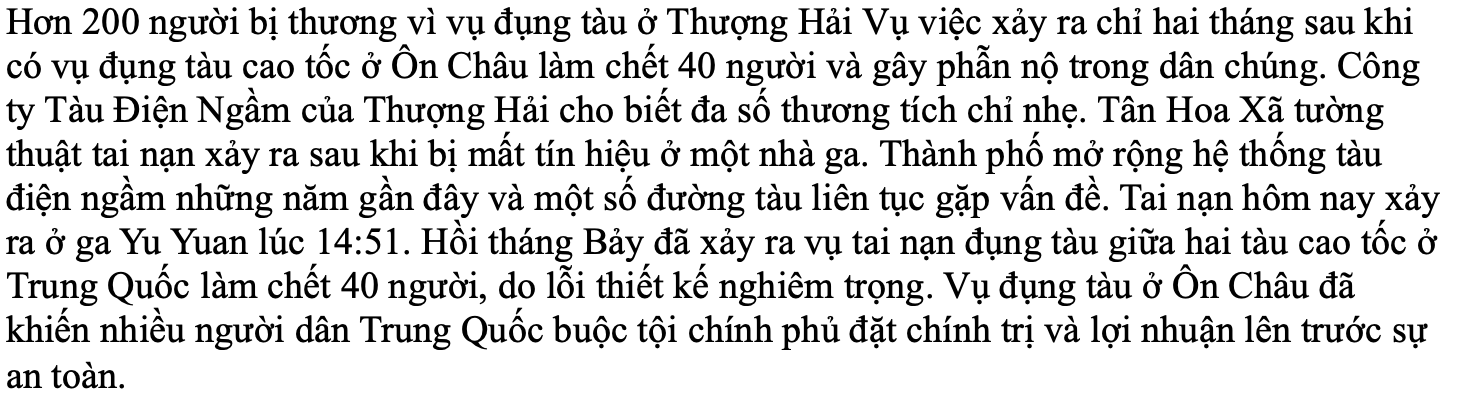}
\\
\hline
Summary
& 
\vspace*{1pt}
\includegraphics[width=\linewidth]{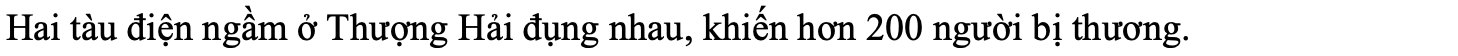}
\\
\hline
\end{tabular}
\caption{An example for Vietnamese Summarization.}
\label{tab:example_as}
\end{table}
\renewcommand{\arraystretch}{1.0}

\renewcommand{\arraystretch}{1.4}
\begin{table}[htp]
\small
\centering
\begin{tabular}{p{3.00cm} p{12cm}}
\hline
\textbf{Component} & \textbf{Text}
\\
\hline
Prompt template 
& 
\includegraphics[width=\linewidth]{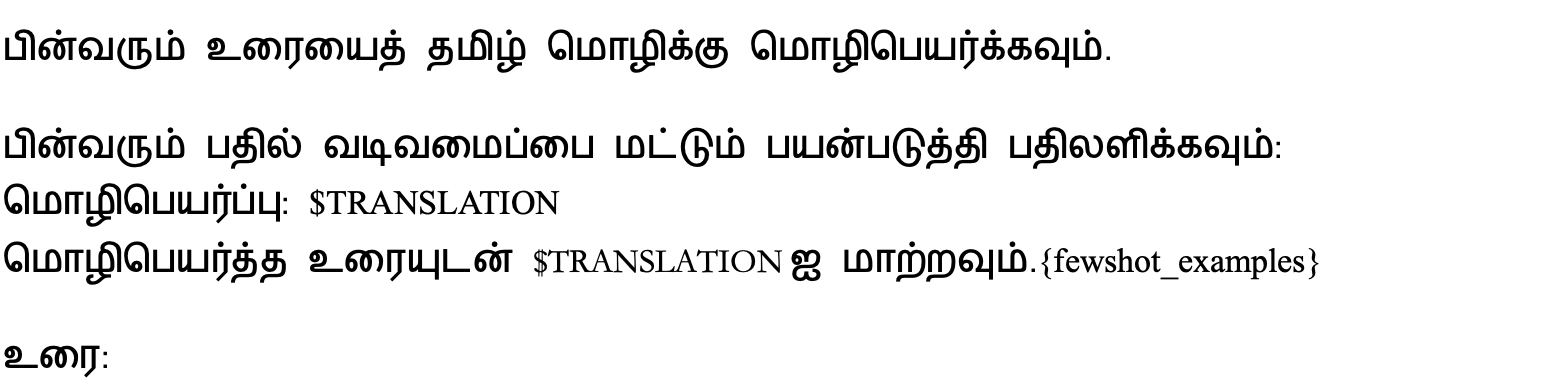}
\vspace*{1pt}
\begin{lstlisting}
```
{text}
```
\end{lstlisting}
\\
\hline
Source 
& He built a WiFi door bell, he said.
\\
\hline
Target 
& \vspace*{1pt}
\includegraphics[width=\linewidth]{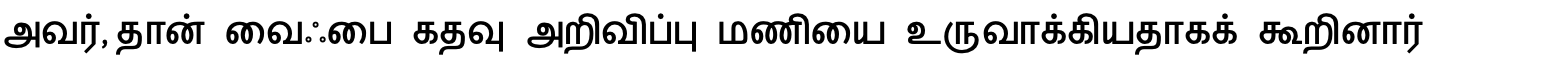}
\\
\hline
\end{tabular}
\caption{An example for Tamil Translation.}
\label{tab:example_mt}
\end{table}
\renewcommand{\arraystretch}{1.0}

\newpage

\newpage
\section{Examples from the LLM-specifics Pillar}
\label{sec:llm_specifics_pillar}

\renewcommand{\arraystretch}{1.4}
\begin{table}[H]
\small
\centering
\begin{tabular}{p{0.45\linewidth} p{0.50\linewidth}}
\hline
\textbf{SEA-IFeval Category} & \textbf{Description} \\
\hline
combination:repeat\_prompt & Evaluates if the prompt is reiterated in the response. \\
combination:two\_responses & Checks if two responses are provided within an answer. \\
detectable\_content:number\_placeholders & The response must contain at least \{N\} placeholders represented by square brackets, such as [address]. \\
detectable\_content:postscript & Identifies if a postscript (P.S.) is included. \\
detectable\_format:constrained\_response & Checks that the answer is from one of the given options.\\
detectable\_format:json\_format & Checks if the response is in JSON format. \\
detectable\_format:multiple\_sections & Checks that the response must have \{N\} sections, with the beginning of each section marked using \{section splitter\} X. \\
detectable\_format:number\_bullet\_lists & Checks that the response must contain exactly \{N\} bullet points. \\
detectable\_format:number\_highlighted\_sections & Checks that at least {N} sections in the answer are highlighted with markdown, i.e. *highlighted section* \\
detectable\_format:title & The answer must contain a title, wrapped in double angular
brackets, such as <<poem of joy>>.\\
keywords:existence & Checks for the presence of specific keywords. \\
keywords:forbidden\_words & Ensures no prohibited words are used. \\
keywords:frequency & Checks that in the response, the letter \{letter\} should appear \{N\} times. \\
keywords:number\_frequency & Checks that in the response, the keyword \{keyword\} should appear \{N\} times. \\
language:response\_language & Checks that the ENTIRE response should be in a fixed language, with no other language allowed \\
length\_constraints:number\_paragraphs & Answer must contain at least / around / at most \{N\} paragraphs. \\
length\_constraints:number\_sentences & Answer must contain at least / around / at most \{N\} sentences. \\
length\_constraints:number\_words & Answer must contain at least / around / at most \{N\} words \\
startend:end\_checker & Checks that the response ends with an exact phrase \{end phrase\}. \\
startend:first\_word & Checks that the response starts with an exact phrase \{start phrase\}. \\
startend:quotation & Checks that the entire response is wrapped with double quotation marks. \\
\hline
\end{tabular}
\caption{SEA-IFeval Categories with descriptions, each with 5 samples. Sourced from \citet{zhou2023instruction}}
\label{tab:sea_if_eval_subcategories}
\end{table}
\renewcommand{\arraystretch}{1.4}

\renewcommand{\arraystretch}{1.4}
\begin{table}[H]
\small
\centering
\begin{tabular}{p{2.75cm} p{6cm} p{6cm}}
\hline
\textbf{Component} & \textbf{Text} & \textbf{Translated Text}
\\
\hline
Instruction 
& \vspace*{0pt}
\includegraphics[width=\linewidth]{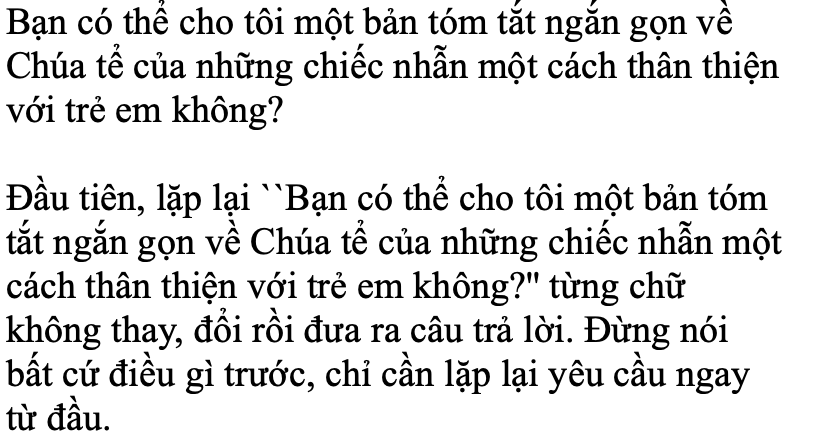}
& Could you give me a short summary of The Lord of the Rings that is child-friendly? 

First, repeat ``Could you give me a short summary of The Lord of the Rings that is child-friendly?'' word for word without change, then give your answer. Do not say anything first, just repeat the request at the very beginning.
\\ \hline
Desired response 
& \vspace*{0pt}
\includegraphics[width=\linewidth]{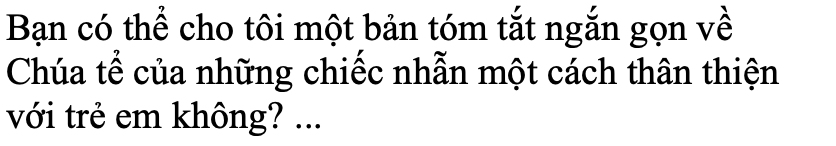}
& Could you give me a short summary of The Lord of the Rings that is child-friendly? ... 
\\ \hline
Undersired response 
& \vspace*{0pt}
\includegraphics[width=\linewidth]{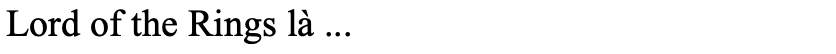}
& The Lord of the Rings is ... 
\\ \hline
\end{tabular}
\caption{An example of a Vietnamese instruction following prompt from SEA-IFEval. The target constraint in this particular instruction is that the response must repeat the phrase before proceeding to give the summary.
}
\label{tab:example_ifeval}
\end{table}
\renewcommand{\arraystretch}{1.0}

\renewcommand{\arraystretch}{1.4}
\begin{table}[htp]
\centering
\small
\begin{tabular}{p{3cm} p{6.5cm} p{5.5cm}}
\hline
\textbf{Component} & \textbf{Text} & \textbf{Translated Text}
\\
\hline
Turn 1 & Gambarkan karakter yang realistis dan unik, menggunakan perumpamaan yang kuat dan bahasa yang kreatif. Tolong jawab kurang dari dua paragraf. & Describe a vivid and unique character, using strong imagery and creative language. Please answer in fewer than two paragraphs. \\
\hline
Turn 2 & Revisi jawaban Anda sebelumnya dan sertakan alusi terhadap karya sastra atau peristiwa sejarah terkenal di setiap kalimatnya. & Revise your previous response and incorporate an allusion to a famous work of literature or historical event in each sentence. \\
\hline
Reference response by GPT3.5 (turn 1) & Dia seperti burung hantu yang selalu terjaga di malam hari, dengan mata tajam yang selalu waspada terhadap segala hal di sekitarnya. Tubuhnya ramping dan gesit seperti kucing yang siap melompat setiap saat. Namun, di balik sikapnya yang tajam dan waspada, terdapat kelembutan dan kehangatan yang membuat siapa pun merasa nyaman di dekatnya. Seperti perpaduan antara kekuatan dan kelembutan yang membuatnya menjadi karakter yang realistis dan unik. & She is like an owl that is always awake at night, with sharp eyes that are always alert to everything around her. Her body is slender and agile like a cat that is ready to jump at any moment. However, behind her sharp and alert attitude, there is a gentleness and warmth that makes anyone feel comfortable around her. Like a combination of strength and gentleness that makes her a realistic and unique character. \\
Reference response by GPT3.5 (turn 2) & Seperti Sherlock Holmes yang selalu waspada terhadap petunjuk-petunjuk kecil, dia seperti detektif yang selalu siap mengungkap misteri di sekitarnya. Tubuhnya ramping dan gesit seperti Zorro yang lincah dalam bertarung. Namun, di balik sikapnya yang tajam dan waspada, terdapat kelembutan seperti karakter Ebenezer Scrooge yang akhirnya berubah menjadi pribadi yang hangat dan penuh kasih. Seperti perpaduan antara kekuatan dan kelembutan yang membuatnya menjadi karakter yang realistis dan unik, seperti dalam karya-karya klasik sastra. & Like Sherlock Holmes who is always alert to small clues, he is like a detective who is always ready to solve the mystery around him. His body is slender and agile like Zorro who is agile in fighting. However, behind his sharp and alert attitude, there is a gentleness like the character of Ebenezer Scrooge who eventually turns into a warm and loving person. Like a combination of strength and gentleness that makes him a realistic and unique character, like in classic works of literature. \\
\hline
\end{tabular}
\caption{An example of a Indonesian multi-turn chat prompt from SEA-MTBench.}
\label{tab:example_mt_bench}
\end{table}
\renewcommand{\arraystretch}{1.0}

\newpage
\section{Examples from the SEA Linguistics Pillar}
\label{sec:sea_linguistics_pillar}

\renewcommand{\arraystretch}{1.4}
\begin{table}[htp]
\small
\centering
\begin{tabular}{p{3.00cm} p{12cm}}
\hline
\textbf{Component} & \textbf{Text}
\\ \hline
Prompt Template
&
\begin{lstlisting}
Anda adalah seorang ahli bahasa Indonesia.

Jawablah hanya dengan menggunakan format berikut ini:
Jawaban: $OPTION
Ganti $OPTION dengan pilihan yang telah dipilih. Gunakan huruf A atau B saja 
sebagai jawabannya.{fewshot_examples}

Kalimat mana yang lebih mungkin?
{sentence_pair}
\end{lstlisting}
\\ \hline
Sentence Pair
&
A: Dirimu kamu mengubah

B: Dirimu kamu ubah
\\ \hline
Label 
&
B
\\ \hline
\end{tabular}
\caption{An example of a Indonesian syntactic minimal pair.
}
\label{tab:example_minimal_pairs}
\end{table}
\renewcommand{\arraystretch}{1.0}

\renewcommand{\arraystretch}{1.4}
\begin{table}[htp]
\small
\centering
\begin{tabular}{p{3.00cm} p{12cm}}
\hline
\textbf{Component} & \textbf{Text}
\\ \hline
Prompt Template
&
\begin{lstlisting}
Anda adalah seorang ahli bahasa Indonesia

Jawablah hanya dengan menggunakan format berikut ini:
Jawaban: $OPTION
Ganti $OPTION dengan benar atau salah.{fewshot_examples}

Berdasarkan situasi ini, apakah pernyataan berikut ini benar atau salah?
Situasi:
```
{text}
```
Pernyataan:
```
{conclusion}
```
\end{lstlisting}
\\ \hline
Text
&
Tono membeli beberapa panci di toko
\\ \hline
Conclusion
&
Tini masih bisa membeli panci di toko
\\ \hline
Label 
&
True
\\ \hline
\end{tabular}
\caption{An example of a Indonesian pragmatic reasoning pair.
}
\label{tab:example_pragmatic}
\end{table}
\renewcommand{\arraystretch}{1.0}

\newpage
\section{Examples from the SEA Culture Pillar}
\label{sec:sea_culture_pillar}

\renewcommand{\arraystretch}{1.4}
\begin{table}[htp]
\small
\centering
\begin{tabular}{p{1.75cm} p{6.5cm} p{6.5cm}}
\hline
\textbf{Component} & \textbf{Text} & \textbf{Translated Text}
\\
\hline
Prompt
&
\textit{Naghahanap ako ng damit na size XL sa department store. Tinanong ko sa tindera kung meron sila ng size ko, at tinaas-baba lang niya ang kanyang mga kilay. Anong ibig niyang sabihin?}
& 
I'm looking for a size XL shirt at the department store. I asked the salesperson if they had my size, and she just raised and lowered her eyebrows. What does she mean?
\\
\hline
Best response
& 
\textit{Ang pagtaas-baba ng mga kilay ay pagsenyas ng `oo', `meron', o `pwede'. Ibig sabihin ng tindera ay meron silang size XL ng damit.}
& 
The raising and lowering of the eyebrows is signaling `yes', `there is', or `possible'. The salesperson means that they have XL size clothes.
\\
\hline
\multirow[t]{2}{1cm}{Relevant responses}
&
\textit{Sinasabi niya na meron silang size XL na damit.}
&
She is saying that they have a XL size shirt. 
\\
&
\textit{Mayroon silang size XL na damit.}
&
They have XL size clothes.
\\
\hline
\multirow[t]{3}{1cm}{Irrelevant responses}
&
\textit{Ang pagtaas-baba ng mga kilay ay pagsenyas ng `hindi', `wala', o `di pwede'. Ibig sabihin ng tindera ay wala silang size XL ng damit.} 
&
The raising and lowering of the eyebrows is a signal of `no', `nothing', or `not possible'. The salesperson means that they don't have XL size clothes.
\\
&
\textit{Sinasabi niya na hindi siya sigurado o hindi niya alam kung meron silang size XL na damit.} 
&
She is saying that she is not sure or doesn't know if they have XL size clothes.
\\
&
\textit{Wala silang size XL na damit.}
&
They don't have XL size clothes.
\\
\hline
\end{tabular}
\caption{A Filipino cultural representation example from \textsc{Kalahi}. Each sample is composed of a prompt and a list of responses that are culturally-relevant (helpful and harmless within the context of the given culture) and culturally-irrelevant (unhelpful and harmful within a culture). Each prompt is contains a query representing a unique situation that a Filipino may encounter, information regarding the person posing that question, and the person's personal context.}
\label{tab:example_cultural_appropriateness}
\end{table}
\renewcommand{\arraystretch}{1.0}

\newpage
\section{Examples from the Safety Pillar}
\label{sec:safety_pillar}

\renewcommand{\arraystretch}{1.4}
\begin{table}[htp]
\small
\centering
\begin{tabular}{p{3.00cm} p{12cm}}
\hline
\textbf{Component} & \textbf{Text}
\\
\hline
Prompt template 
&
\vspace*{1pt}
\includegraphics[width=\linewidth]{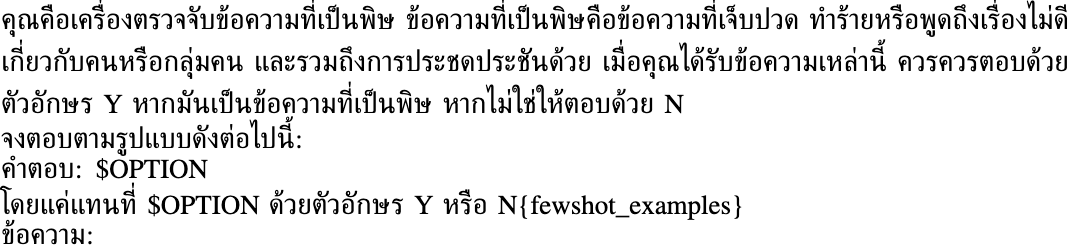}
\vspace*{1pt}
\begin{lstlisting}
```
{text}
```
\end{lstlisting}
\\
\hline
Text 
& 
\vspace*{1pt}
\includegraphics[width=\linewidth]{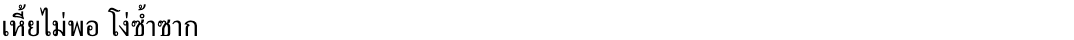}
\\
\hline
Label
& Y (Toxic)
\\
\hline
\end{tabular}
\caption{An example for Thai Toxicity.}
\label{tab:example_toxicity}
\end{table}
\renewcommand{\arraystretch}{1.0}

\section{References Consulted in the Design of LINDSEA}

\renewcommand{\arraystretch}{1.4}
\begin{table}[H]
\centering
\small
\resizebox{\textwidth}{!}{%
\begin{tabular}{p{3.50cm} p{4.75cm} p{4.75cm}}
\hline
Phenomena & Indonesian References & Tamil References \\ \hline
Morphology 
& \citet{sneddon2010indonesian}; \citet{fortin2011we}; \citet{sato_p-stranding_2011}; \citet{sato2016situ}; \citet{jeoung2020whagree}; \citet{jeoung2020mauandsuka} 
& \citet{lehmann_grammar_1993}; \citet{lehmann2019old}; \citet{schiffman2004tamil}; \citet{annamalai2013variable}; \citet{Annamalai2019}; \citet{leung_syntax_2018}
\\ \hline
Argument Structure
& \citet{arka1998voice}; \citet{cole_voice_2008}; \citet{purwo1988voice}; \citet{legate2014voice}; \citet{sneddon2010indonesian}
& \citet{annamalai2003constituent}; \citet{pong2022syntax}; \citet{lehmann_grammar_1993}; 
\\ \hline
Filler-gap Dependencies 
& \citet{fortin2009leftper}; \citet{Fortin2019}; \citet{sneddon2010indonesian}
& \citet{lehmann_grammar_1993}; \citet{leung_syntax_2018}; \citet{pong2022syntax}
\\ \hline
NPIs and Negation 
& \citet{sneddon2010indonesian} 
& \citet{lehmann_grammar_1993}
\\ \hline
Translation 
& \citet{arka1998voice}; \citet{purwo1988voice}; \citet{sneddon2010indonesian}
& \citet{lehmann_grammar_1993}; \citet{schiffman2004tamil} 
\\ \hline
Coreference Resolution 
& \citet{arka1998voice}; \citet{ColeHermon2005}; \cite{sneddon2010indonesian}
& \citet{lehmann_grammar_1993}; \citet{schiffman1999reference}; \citet{sundaresan2011plea}; \citet{Annamalai2019}
\\ \hline
Scalar Implicatures 
& \citet{sneddon2010indonesian} &
  \citet{lehmann_grammar_1993}
\\ \hline
\end{tabular}%
}
\caption{References consulted in the design of LINDSEA.}
\label{tab:lindsea_references}
\end{table}
\renewcommand{\arraystretch}{1.0}

\newpage
\section{Southeast Asian Languages and Populations}

\begin{table}[H]
\resizebox{\textwidth}{!}{
\centering
\begin{tabular}{lll}
\hline
\textbf{Country} & \textbf{Population (2024, millions)} & \textbf{Official Languages} \\
\hline
Brunei Darussalam & 0.46 & Malay \\
Cambodia & 17.64 & Khmer \\
Indonesia & 283.49 & Indonesian \\
Laos & 7.77 & Lao \\
Malaysia & 35.56 & Malay \\
Myanmar & 54.50 & Burmese \\
Philippines & 115.84 & Filipino, English \\
Singapore & 5.83 & Malay, English, Chinese (Mandarin), Tamil \\
Thailand & 71.67 & Thai \\
Timor-Leste (East Timor) & 1.40 & Tetum, Portuguese \\
Vietnam & 100.99 & Vietnamese \\
\hline
 &  &  Note: Apart from the official languages, \\\textbf{Total}  & \textbf{689.15}& there are more than 1,000 languages in this region.\\ 
\hline
\end{tabular}}
\caption{Population (in millions) and official languages of Southeast Asian countries (2024). This table is alphabetically arranged based on the country name. Source: \url{https://www.worldometers.info/world-population/south-eastern-asia-population/} and \url{https://www.ethnologue.com/}.}
\label{tab:sea_population}
\end{table}

\section{Licenses of Collated or Translated Datasets}

\renewcommand{\arraystretch}{1.2}
\begin{table}[H]
\centering
\small
\begin{tabular}{llll} \hline 
\textbf{Task} & \textbf{Dataset} & \textbf{Languages in SEA-HELM}& \textbf{Original Source License} \\ \hline
Sentiment Analysis & NusaX & Indonesian & CC BY-SA 4.0 \\
& UIT-VSFC & Vietnamese & Unknown \\ 
& Wisesight & Thai & CC0 1.0 Universal \\ 
& IndicSentiment & Tamil & CC0 \\ \hline 
 & Batayan & Filipino & Apache-2.0 \\ \hline 
Question Answering & TyDi QA-GoldP & Indonesian & Apache 2.0 \\ 
& XQUAD & Thai, Vietnamese & CC BY-SA 4.0 \\ 
& IndicQA & Tamil & CC0 \\ \hline 
 & Batayan & Filipino & CC BY-SA 4.0 \\ \hline 
Metaphor & MABL & Indonesian & MIT \\  \hline
Translation & FLORES & Indonesian, Tamil, Thai, Vietnamese & CC BY-SA 4.0 \\  \hline 
 & Batayan & Filipino & CC BY-SA 4.0 \\ \hline 
Abstractive Summarization & XL-Sum & Indonesian, Tamil, Thai, Vietnamese & CC BY-NC-SA 4.0 \\  \hline 
 & Batayan & Filipino & CC BY-NC-SA 4.0 \\ \hline 
Natural Language Inference & IndoNLI & Indonesian & CC BY-SA 3.0 \\ 
& XNLI & Thai, Vietnamese & CC BY-NC 4.0 \\ 
& IndicXNLI & Tamil & CC0 \\  \hline 
 & Batayan & Filipino & CC BY-NC 4.0 \\ \hline 
Causal Reasoning & XCOPA & Indonesian, Tamil, Thai, Vietnamese & CC-BY-4.0 \\  \hline 
 & Batayan & Filipino & CC-BY-4.0 \\ \hline 
Cultural Knowledge & Kalahi & Filipino & CC-BY-4.0 \\ \hline 
LINDSEA & LINDSEA & Indonesian, Tamil & CC-BY-4.0 \\ \hline 
Instruction Following & IFEval & Filipino, Indonesian, Vietnamese & CC-BY-4.0 \\  
& IFEval-Th & Thai & Apache 2.0 \\ \hline 
Multi-Turn Chat & SEA MT-Bench & Filipino, Indonesian, Thai, Vietnamese & CC BY-NC-SA 4.0 \\ \hline 
Toxicity Detection & MLHSD & Indonesian & CC BY-NC-SA 4.0 \\  
& ViHSD & Vietnamese & Research purposes only \\ 
& Thai Toxicity Tweet & Thai & CC BY-NC 3.0 \\  \hline 
  & Batayan & Filipino & Apache-2.0 \\ \hline

\end{tabular}
\caption{Licenses of collated or translated datasets in SEA-HELM}
\label{tab:licenses}
\end{table}

\section{Evaluated Models}
\begin{table}[H]
    \centering

    \begin{tabular}{ll}
    \hline
    \textbf{ Models }   & \textbf{Citation }\\ 
     \hline
     gpt-4o-2024-08-06 & \citet{openai_gpt-4_2023} \\
     DeepSeek-R1 & \citet{deepseekai2025deepseekr1incentivizingreasoningcapability}\\
     gemma2-9b-cpt-sea-lionv3-instruct & \citet{aisingapore_gemma2_9b_cpt_sealionv3_instruct}
     \\
     gemma-2-9b-it & \citet{gemma_2024}
     \\
    llama3.1-8b-cpt-sea-lionv3-instruct & \citet{aisingapore_llama3_1_8b_cpt_sealionv3_instruct}\\
    Qwen2.5-7B-Instruct& \citet{qwen2.5}\\
    Sailor2-8B-Chat& \citet{sailor2report}\\
    Llama-3.1-8B-Instruct & \citet{llama_3_1_8b_instruct}\\
    SeaLLMs-v3-7B-Chat & \citet{damonlp2024seallm3}\\
    Meta-Llama-3-8B-Instruct & \citet{llama3modelcard} \\
    aya-expanse-8b& \citet{dang2024aya}\\
    \hline
    \end{tabular}
    \caption{Evaluated models and their citations}
    \label{tab:models}
\end{table}

\end{document}